\documentclass[conference]{IEEEtran}
\usepackage{cite}
\usepackage{amsmath,amssymb,amsfonts}
\usepackage{algorithmic}
\usepackage{graphicx}
\usepackage{subfig}
\usepackage{float}
\usepackage{flushend}
\usepackage{placeins}
\usepackage{textcomp}
\usepackage{xcolor}
\usepackage{caption}

\def\BibTeX{{\rm B\kern-.05em{\sc i\kern-.025em b}\kern-.08em
    T\kern-.1667em\lower.7ex\hbox{E}\kern-.125emX}}
\begin{document}

\title{Domain Adaptation of Reinforcement Learning Agents based on Network Service Proximity
}

\author{\IEEEauthorblockN{Kaushik Dey, Satheesh K. Perepu}
\IEEEauthorblockA{\textit{Ericsson Research (Artificial Intelligence)} \\
India\\
\{deykaushik,perepu.satheesh.kumar\}@ericsson.com}
\and
\IEEEauthorblockN{Pallab Dasgupta, Abir Das}
\IEEEauthorblockA{\textit{Indian Institute of Technology, Kharagpur} \\
India \\
\{pallab,abir\}@cse.iitkgp.ac.in}
}

\maketitle

\begin{abstract}
The dynamic and evolutionary nature of service requirements in wireless networks has motivated the telecom industry to consider intelligent self-adapting Reinforcement Learning (RL) agents for controlling the growing portfolio of network services. Infusion of many new types of services is anticipated with future adoption of 6G networks, and sometimes these services will be defined by applications that are external to the network. An RL agent trained for managing the needs of a specific service type may not be ideal for managing a different service type without domain adaptation. We provide a simple heuristic for evaluating a measure of proximity between a new service and existing services, and show that the RL agent of the most proximal service rapidly adapts to the new service type through a well defined process of domain adaptation. Our approach enables a trained source policy to adapt to new situations with changed dynamics without retraining a new policy, thereby achieving significant computing and cost-effectiveness. Such domain adaptation techniques may soon provide a foundation for more generalized RL-based service management under the face of rapidly evolving service types.
\end{abstract}


\section{Introduction}
The infusion of automated intelligent control in service management of large wireless networks is becoming ubiquitous and is widely anticipated to become the key enabler for managing the large milieu of service types to be supported in future 6G networks. In present 5G networks, services are broadly bundled into a few types such as {\em Conversational Video} (CV), {\em Ultra Reliable Low Latency Communication} (URLLC), and {\em massive IoT} (mIoT), but the number of such types is expected to multiply considerably in near future.

Existing literature and practices have demonstrated the benefits of using Reinforcement Learning (RL) agents to control the network service parameters in an optimal manner~\cite{perepu2022multi,paper:reinforcement2,paper:reinforcement3}. An RL agent may be trained for a specific service and radio environment to optimally control the parameters in air interface or core networks, such that expectations and demands on the network can be fulfilled autonomously. Such an RL agent may sit in the orchestration layer, core network layer, or even in edge nodes like gNodeBs\footnote{gNodeBs in 5G are the functional equivalent of base stations in traditional systems, and are responsible for radio communication with user equipment (UE) in its coverage area, namely cell site.}. This work assumes that the RL policies reside in the core or orchestration layer, and use a single policy to manage a cluster of gNodeBs. However, such a method alone may not be sufficient, as in the near future the journey from 5G to 6G will escalate the need to have AI native functions across all parts of the network~\cite{paper:6G,paper:6G_AI}. Subsequently, external applications will dynamically spawn new service types and consumption patterns in the network will change. The underlying radio environment will also change with changing mobility patterns and associated handovers across gNodeBs~\cite{paper:6G_network}. In an urban scenario, the radio environment may change rapidly (say, every 10 minutes, or less).

Hence the RL policy may need frequent retraining. But from energy and compute considerations, it will not be viable to {\em retrain} a RL agent despite change in consuming service type or traffic distributions. On the other hand, an agent trained in a certain environment, such as certain {\em user equipment} (UE) distribution across gnodeBs, may not perform optimally in a different environment with a changed UE distribution . Hence it will be imperative to have AI algorithms that can enable a RL agent to adapt to different  types of services and environments. This ability to adapt to different domains, new services, or environment conditions with limited or no re-training may be the key competitive differentiation for effective deployment of intelligent agents in service management of future networks~\cite{paper:6G_network}. In RL parlance, we wish to be able to adapt a policy learned in one domain by corresponding observations and actions in another related domain such that our next state observations are similar or close enough~\cite{paper:zhang_cycle}.

In this paper, we leverage domain adaptation techniques studied in RL literature to prepare a method for adapting a RL agent trained for one kind of service to another kind of service. We demonstrate that the speed and quality of adaptation relies on a measure of proximity between the original and the target services. This leads us to a heuristic for choosing the source policy that can be quickly and reliably adapted to the target service. The main contributions of the paper are:
\begin{enumerate} 

    \item We provide a method for RL policy generalization across services in the telecom domain  using {\em Cycle-consistent Generative Adversarial Networks} (Cycle-GANs), for the first time. Results demonstrate our method is 5-6 times sample efficient (Figure \ref{fig:URLLC_miot}, \ref{fig:miot_urllc}) than existing methods
    
    \item We validate the proposed method for situations where the radio environment changes very frequently due to mobility patterns. The proposed method enables quick adaptation which is affordable from compute considerations and sample efficiency perspective. Our method outperforms retraining by almost 8-9 times (Table~\ref{tab:dist_training}) in sample efficiency and enables policy adaptation in zero-touch settings as reward signals are not needed in the new environment.
    
    \item We propose a heuristic that gives guidance on how well domain adaptation would perform before initiating the correspondence mapping. Domain adaptation will not work well in every situation and our suggested heuristic successfully identifies such distribution changes apriori.  Additionally, such a metric will provide guidance regarding which service to consider as a source such that the same can be adapted to a wide number of new services. 

\end{enumerate}
The paper is organized as follows. Section~\ref{sec:domain_adapt} outlines related work in the domain adaptation area. Section III highlights the need for domain adaptation and presents the proposed methodology. It also elaborates on the proposed heuristics for estimating service proximity. Section~\ref{sec:exper_setup} outlines the experimental setup and Section~\ref{sec:results} presents the results obtained using the proposed approach. Concluding remarks and future directions are provided in Section~\ref{sec:conc}. 

\section{Related Work on Domain Adaptation for RL}\label{sec:domain_adapt}
In the context of RL, existing literature provides various approaches for adapting a policy to a new domain. These may be broadly classified into the following three approaches:
\begin{enumerate}

    \item {\em Representational Learning.} In order to align between two domains, this method attempts to find an invariant representation with respect to the source and target domains~\cite{paper:representive1,gupta2017learning,paper:representative2}. The approach is based on the assumption that a policy, which is learned on basis of the commonality of the two environments, is unaffected by the changes in the target domain with respect to the source. However, this method faces a challenge when the difference between two domains cannot be accounted for by creating representational augmentations. Also when the dynamics of the two environments differ, as it is in our case with three different services, this technique requires paired data which is both expensive and time-consuming to collect.

    \item {\em Meta Learning}. Here the model is trained over a distribution of tasks~\cite{finn2017model}, such that it can leverage its knowledge for learning a new task from the same distribution. Deep meta-reinforcement learning~\cite{wang2016learning} extends Meta-Learning to a RL setting where a model learns to adapt to a new task by learning a policy to select an optimal action in the new environment. Such approaches have proven to be useful to learn a near-optimal trajectory of actions when exposed to a new environment but with fewer samples. However such techniques often excel when the model is trained with a distribution of tasks rather than a single task. Hence this may not be particularly suitable in a one-to-one transfer, which we attempt in this work.
 
    \item {\em Learning Correspondences.} In contrast to learning invariant representations, here we concentrate on learning the mapping between states and aligning actions such that next-state transitions are correspondent between the domains. Such a method~\cite{paper:zhang_cycle} can be implemented using Cycle-GANs~\cite{paper:cyclegan} and does not need paired state-action tuples across environments. Most importantly knowledge of the reward function in the target environment is not required. The method in~\cite{paper:zhang_cycle} relies on learning correspondences across input modalities, physics parameters, and different morphologies. For our experiments, the nature of packet transmission varies across all services, and also the {\em key performance indicator} (KPI) varies between the QoE and URLLC/mIoT services. Hence the change in any of the control parameters, {\em packet priority}, or {\em maximum bit rate} (MBR) affects the KPI differently.  Therefore principally we are dealing with a situation where physics parameters or dynamics are altered between environments and the adapted policy must find a near-optimal state-action correspondence in the new environment. 

\end{enumerate}
Cycle-GANs that use cycle-consistency loss have traditionally been used to perform image-to-image translations and reconstructions. However, Cycle-GANs have also been applied to domain adaptation in images~\cite{paper:cycle_gan3} and subsequently for sim-to-real challenges by using image translations between simulation and real-world images~\cite{paper:cycle_gan5}. More recently Cycle GANs have been explored for reinforcement learning whereby RL-scene consistency loss~\cite{paper:cycle_gan4} is used to learn a task-aware transition that is invariant with respect to Q values of images.  

All the above mentions used Cycle-GANs primarily for vision correspondences till \cite{paper:zhang_cycle} used it for aligning different dynamics and cross-physics alignments. However, to this date, such methods have not been tried in the telecom or network domain where the nature of packet transmissions varies over different services and even the expectations for each service may be measured through different KPIs.

In a multi-dimensional action setting, the time period of a single step of an episode may depend on the type of action taken, which makes establishing correspondences more challenging. In our case, a priority action needs about 40 seconds to have a stabilizing effect in the simulator while the MBR action is visible in less than 2 seconds. It also needs to be considered that in a network setting, data collection (samples of 3 element tuples) is also an expensive process as compared to a MUJOCO~\cite{paper:mujoco} or visual simulation environment~\cite{paper:visual_RL} and hence the correspondence needs to be established over a much smaller set of data and the proposed approach needs to prove it's sample efficiency of over existing techniques. 

\section{Methodology Overview}
We look at the following situations where domain adaptation may be essential. The RL agent uses {\em Priority} and {\em Maximum Bit Rate} (MBR) as control parameters. 
\begin{enumerate}
 
    \item {\em Mobility patterns change during the day.} Therefore, the {\em User Equipment} (UE) distributions across gnodeBs also change, which alters the traffic pattern. 
    
    \item {\em New services may be introduced.} In our study, we look at three different types of services, namely Conversational Video (CV), Ultra Reliable Low Latency Communication (URLLC), and massive IoT (mIoT). Initially, we train a policy using only one service, for example, URLLC. Then we attempt to adapt the URLLC policy to other service types, such as mIoT or CV.
    
\end{enumerate}
 In our work, we propose a method whereby, instead of retraining, we build correspondences between state-action pairs across two environments. In the process of building correspondences, we do not use any reward signal from the new environments.  Such adaptions are faster, computationally cheaper, and hence more affordable to be deployed in edge devices with low computing power. Additionally, the process can be easily automated without human interference as manual reward engineering is not needed.
 
For the purposes of this work, we have used a Network Emulator which can simulate three types of services (i) Conversational Video (CV), (ii) ultra-reliable low latency communications (URLLC), and (iii) Massive IoT (mIoT). Here we can set an intent or expectation on a specific {\em key performance indicator} (KPI) for each service. The expectation for the URLLC and mIoT services is set in terms of the KPI, {\em Packet Error Rate} (PER), while the CV service is measured by the KPI {\em Quality of Experience} (QoE). For tractability, we have chosen two control parameters that the algorithm can modify, namely, {\em Maximum Bit Rate} (MBR) and {\em Packet priority}. 

\subsection{The need for domain adaptation}
\begin{figure}
    \centering
    \includegraphics[scale=0.4]{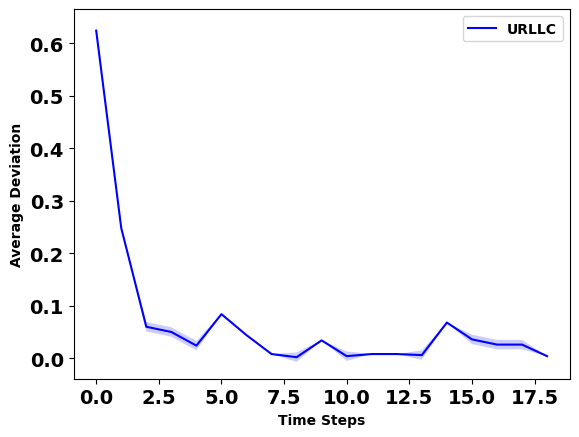}
    \caption{Performance of URLLC training on the URLLC service}
    \label{fig:URLLC_exec}
\end{figure}

\begin{figure}
    \centering
    \includegraphics[scale=0.4]{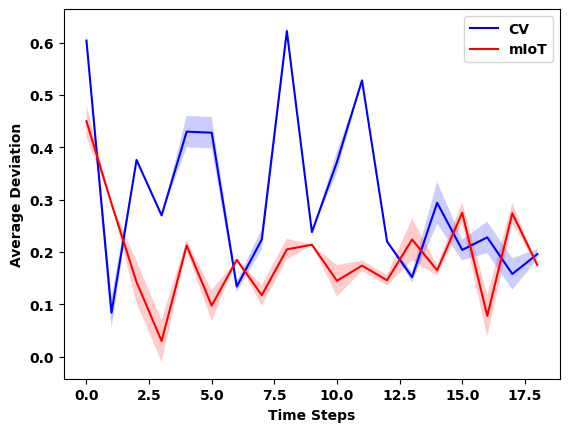}
    \caption{Performance of URLLC policy on mIoT and CV}
    \label{fig:URLLC_exec_m_c}
\end{figure}

For the purpose of our studies, we have used a reinforcement learning (RL) technique named PPO \cite{paper:ppo} to generate an optimal policy in each source domain and apply it as is in a target domain. For example, the source domain can be URLLC and the target domain can be mIoT or CV. Figure~\ref{fig:URLLC_exec} and Figure~\ref{fig:URLLC_exec_m_c} corresponds to an experiment where a URLLC policy is tested on mIoT and CV environments. It is very evident that the URLLC policy performs quite below par and is not able to attain the desired goal set for either mIoT or CV.

Likewise, a URLLC policy trained in one environment with certain UE distribution across a few gNodeBs under-performs when the UE distribution changes. Figure~\ref{fig:modified_distr_1} shows a situation where the distribution of UEs across gNodeBs changes from Gaussian distribution to uniform due to mobility.

Hence in this paper, we propose to build a domain adaptation technique by virtue of which a policy trained in one environment can be adapted and aligned to another environment or another service.

\subsection{Domain adaptation methodology} \label{sec:domainadapt}
For cross-domain alignment, we propose a method that learns correspondences across the unordered state and action pairs. We train a combination of state correspondence and action correspondence models along with dynamics cycle-consistency~\cite{paper:zhang_cycle}, which aligns the sequence of the next states across domains. We use a tuple consisting of the current state, action, and next state in both source and target domains. The policy obtained during training for the source environment is used to collect batches of tuples across both domains. The unordered pairs of tuples are then aligned through Cycle-consistent Generative Adversarial Networks or Cycle-GANs~\cite{paper:cyclegan,paper:cycle_gan2,paper:cycle_gan3}. 

We describe the methodology through an use case where a policy is trained on the URLLC service and domain-adapted on the mIoT service. 

\begin{figure*}
    \centering
    \includegraphics[scale=0.45]{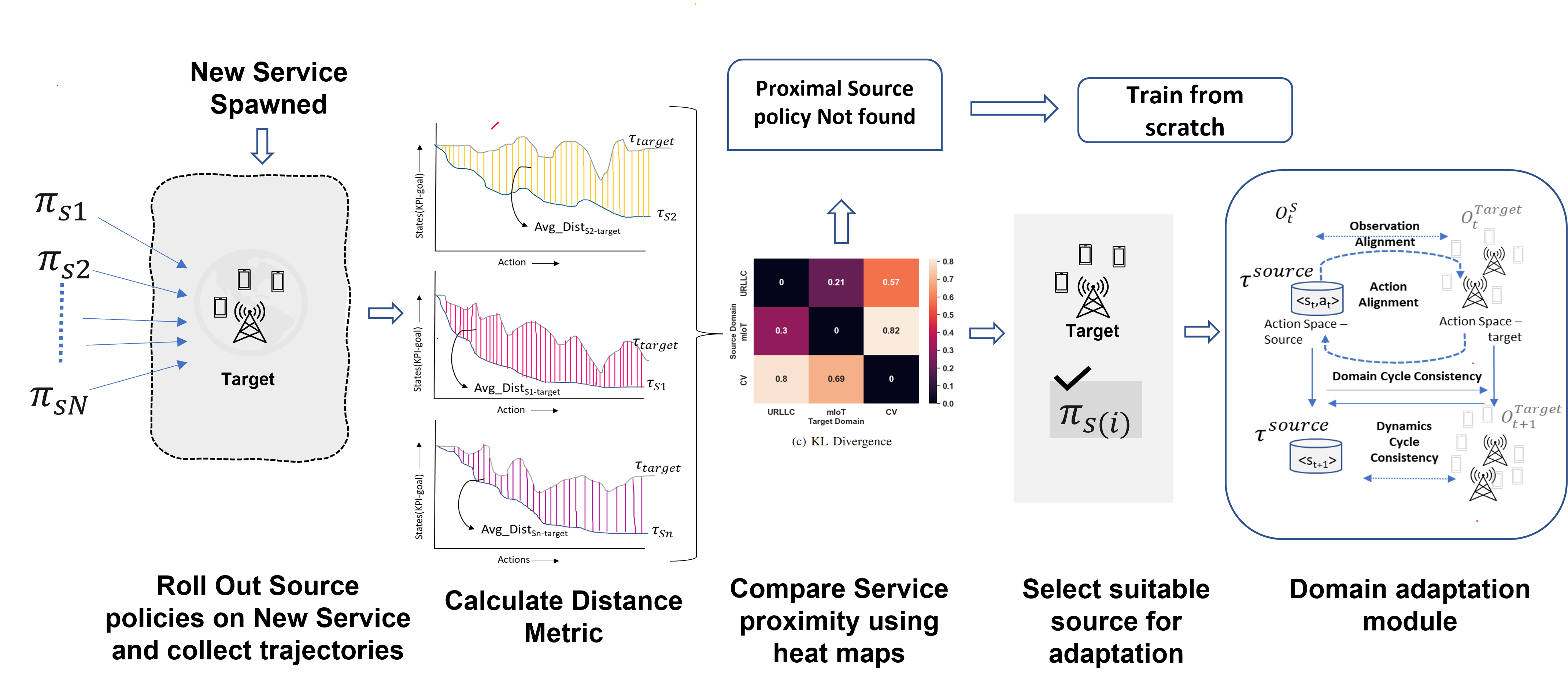}
    \caption{Overview of the proposed domain adaptation methodology}
    \label{fig:domain_adaptation}
\end{figure*}
\begin{enumerate}

    \item The trained URLLC policy is executed in the URLLC environment for multiple episodes to collect 3-tuples of the form (state, action, next-state):
    \[ \tau^u = ( s_t^u, a_t^u, s_{t+1}^u ) \]

    \item The trained URLLC policy is also executed in the mIoT environment, without any adaption, to collect 3-tuples of the form: 
    \[ \tau^m = ( s_t^m, a_t^m, s_{t+1}^m ) \]
    
    \item In the next step, in order to align and build correspondences across state-action pairs we create an {\em Observation Alignment Function}, $G$: $S^{u}\rightarrow S^{m}$, and an {\em Action Alignment Function}, $H$: $S^{u} \times A^u \rightarrow A^{m}$ and its inverse, $H^{-1}$: $S^{m}\times A^m \rightarrow A^u$. Hence given initial correspondent states across URLLC and mIot, the actions are chosen in such a way that the next states also remain correspondent, that is, if $S^{u}_t \leftrightarrow S^{m}_t$ and $S^{u}_{t+1} \leftrightarrow S^{m}_{t+1}$, then $(s_t^u,a_t^u) \leftrightarrow (s_t^m,a_t^m)$, which we refer to as {\em action correspondence}. The mapping of states across domains is achieved using adversarial training and action correspondences are achieved through cycle-consistency loss~\cite{zhu2017unpaired}, such that the action obtained through $H$ can be translated back through $H^{-1}$.
    
    \item A domain cycle consistency loss, $Loss_{domclc}$, in introduced whereby $H^{S^{u},A^{u}}$ is used to obtain $\tilde{a}^m$. Then $H^{-1}(S^m,\tilde{a}^m)$ is applied to get $\tilde{a}^u$ which is expected to be equal to $A^u$.
    
    \item In addition to the state and action correspondence we also need the dynamics correspondence which is introduced through a {\em dynamics cycle-consistency function} for mapping the transition dynamics across the two domains. Given the $S^t_m$ and $A^t_m$ obtained from $G$ and $H$ respectively, we can create a mapping, such that, $T_m(\Tilde{s}^t_m,\Tilde{a}^t_m) \leftrightarrow s^{t+1}_u$, where $T_m$ is the transition function in the mIoT domain such that $T_m(s^t_m,a^t_m)\rightarrow s^{t+1}_m$. Therefore $T_m(\Tilde{s}^t_m,\Tilde{a}^t_m)$ should map to $G:(s^{t+1}_u)$, which can also be expressed as a cycle-consistency function.
    
    \item The above objectives are optimized together in a loss function \cite{paper:zhang_cycle} expressed as:
    {\small
    \begin{eqnarray*}
    Loss_{full} &=& \lambda_{0}(Loss_{dyncyc}(G,H)) + \\
    & & \lambda_{1}(Loss_{adv}(H,D_A^m) + Loss_{adv}(H^{-1},D_A^u)\\ 
    & & \qquad + Loss_{domclc}(H, H^{-1})) + \\
    & & \lambda_{2}(Loss_{adv}(G,D_{S_m}))
    \end{eqnarray*}
    }
    where $\lambda_0$, $\lambda_1$ and $\lambda_2$ are constants to balance the losses.
    \item The above steps performed over a much-reduced sample space (as compared to training from scratch) provide us with a domain-adapted policy on the mIoT domain from the original URLLC policy.
    
\begin{figure*}
    \centering
     \subfloat[Euclidean]{\includegraphics[width=5cm]{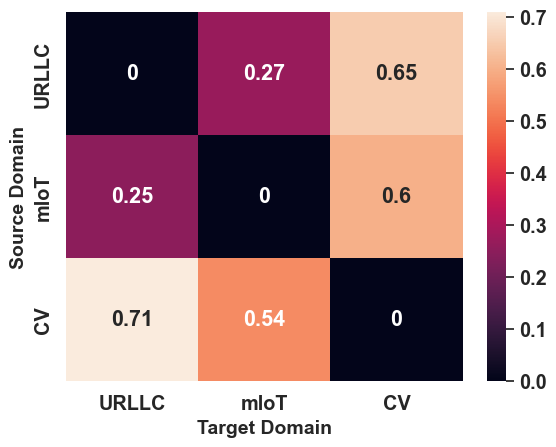}\label{fig:HM_E} $\qquad$}
     \subfloat[Manhattan]{\includegraphics[width=5cm]{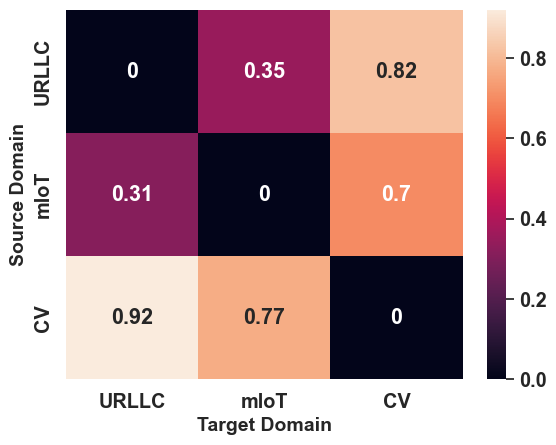}\label{fig:HM_M} $\qquad$}
     \subfloat[KL Divergence]{\includegraphics[width=5cm]{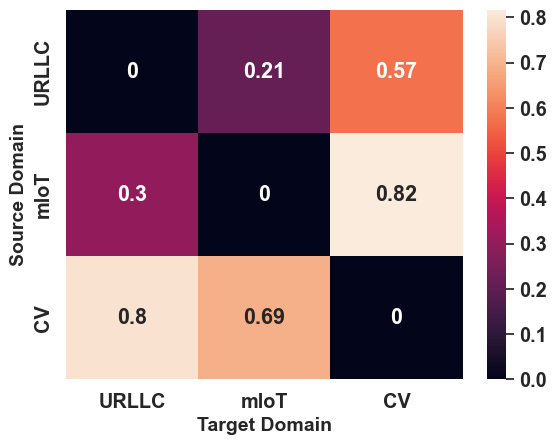}\label{fig:H_K}}
    \caption{Heat maps between the performance on source domain vs target domain}
\end{figure*}

\end{enumerate}

An overview of the domain-adaption methodology is shown in Figure~\ref{fig:domain_adaptation}. $S^{source(1..n)}$ denotes the source states for each of which a trained policy $\pi_{s(1...N)}$ is available. $S^{target}$ denotes the states of the new/changed domain which is spawned either for a new service or due to modified radio environment arising out of mobility and handover changes across gNodeBs. Here the same method is applied whereby a suitable source $S^{source(i)}$ is first selected by Service Proximity heuristic and then  the observation alignment function is applied as $G:S^{source_(i)}\rightarrow S^{target}$, action alignment function is $H:S^{source_(i)} \times A^{source_(i)} \rightarrow A^{target}$, and its inverse is $H^{-1}:S^{target}\times A^{target} \rightarrow A^{source_(i)}$. The rest of the domain adaptation process is similar to the one detailed above.    

\subsection{Heuristics for Service Proximality} \label{sec:heuristic}
Our studies on domain adaptation are presented in Section~\ref{sec:results}. We observed that choosing the source policy is an important step in achieving a better policy in less time. In the presence of a milieu of existing services, it is therefore important to choose the source policy in a judicious way for adapting to a new service. We propose a heuristic for making this choice based on {\em service proximity}.

If two domains are similar, there is a high chance that the same action is best in similar states. The distance between the similar states in the two domains also will be less, and we may use this distance as a heuristic. In order to calculate the distance between similar states across two domains, the concept of lax-bisimulation distance~\cite{NIPS2008_6602294b} has been widely used. Closely related to bisimulation is the theory of MDP homomorphism which helps to formulate a state-dependent action mapping~\cite{paper:homomorphism1,paper:homomorphism2}. Both have been extensively used in transfer learning. However, the lax-bisimulation metric needs information on reward in both environments along with Kantarovich distance between the transition probabilities. Both of these metrics are often not available a priori for the new domain. Additionally, reward formulation may need a reward engineering effort which is often manual and time-consuming. Hence none of these approaches are suitable as we assume {\em no information is available for the new domain}. 

Our heuristic is computed by piloting a source policy on a target domain. We consider the sequence of actions taken by a source policy, from an initial state, $S_0$, and record the trajectory of states, in form of $deviation_A=\tau_A-Target_A$, visited in the source domain, $A$. We then apply the same sequence of actions from $S_0$ in the target domain and record the trajectory of states, $deviation_B=\tau_B-Target_B$, visited in the target domain. Thereafter we compute the {\em distance} between $deviation_A$ and $deviation_B$. The $deviation$ metric adopted here is in accordance with the state definition for goal-conditioned factors as highlighted in Section~\ref{sec:results}. The proximity between the source and target domains is determined in terms of averaged distances over multiple trajectories. Here the multiple trajectories correspond to different initial states $S_0$ in both domains and/or different target values in different domains.

We compute three types of distances, namely {\em Euclidean}, {\em Manhattan}, and {\em KL-divergence}, between the target domain and each of the source domains. The heat maps of the Euclidean distances, Manhattan distances, and KL-divergence are shown in Fig~\ref{fig:HM_E}, Fig~\ref{fig:HM_M}, and Fig~\ref{fig:H_K} respectively. In order to validate our prescription, we have considered the known services, URLLC, mIoT, and CV, for which we have adequate data and trained policies. For computing the heat maps we consider all ordered pairs of these services, each time treating one of these as the source and the other as the target.

From these figures, it is evident that the heuristic distance is smaller for $\langle$ URLLC, mIoT $\rangle$ and $\langle$ mIoT, URLLC $\rangle$, as compared to the ordered pairs involving CV. This is in agreement with domain knowledge and further corroborated by the results presented in Section~\ref{sec:results}.

\section{Experimental Setup}\label{sec:exper_setup}
\begin{figure}
    \centering
    \includegraphics[scale=0.25]{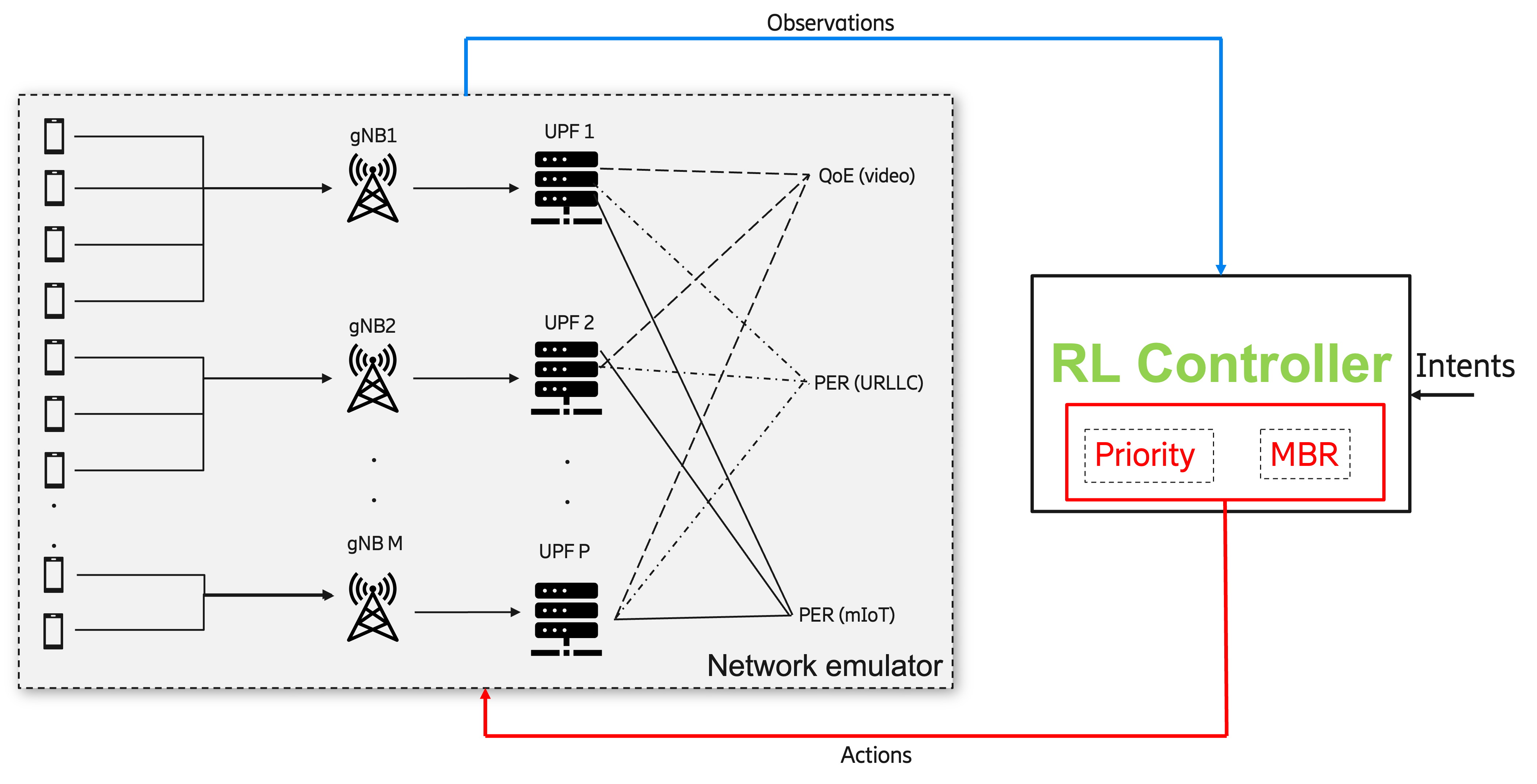}
    \caption{Overview of the Network Emulator}
    \label{fig:network_emulator}
\end{figure}

In order to test the domain adaptation on telecom services, a custom emulator has been built to depict the working of the 5G network. The overview of the network emulator is shown in Figure~\ref{fig:network_emulator}. The emulator supports three services, namely {\em CV}, {\em URLLC}, and {\em mIoT}. These services generate different types of traffic patterns as the traffic they carry is different from one another. In the emulator, UE's are modeled to send/receive the traffic via UPFs and gNodeBs to/from the application layer. For simplicity and clarity on inferences drawn from the results, we have assumed that at any time, UE uses only one service. 

To reiterate, we chose one KPI for each one of the services listed above, namely {\em QoE} for CV, and {\em PER} for URLLC and mIoT. Two variables, namely {\em priority} and {\em maximum bit rate} (MBR), are chosen to control these KPIs. Both the priority and MBR values are directly proportional to the amount of downlink throughput that UE's can get. 

In this work, we chose to control priority at the service level, that is, the priority value chosen for the service is applied to all the UE's using the service. However, we decided to control MBR at the UE level. The range of values considered for these variables is shown in Table~\ref{tab:spaces}. 

\begin{table}[ht]
    \centering
    \caption{Ranges chosen for KPIs}
    \begin{tabular}{ll}
     \textbf{\sf KPI}    &  \textbf{\sf Range}\\
     \hline \hline
      QoE   & $[1,5]$ \\ \hline
      Packet Loss & $[0\%,100\%]$ \\ \hline
      Priority & $\{0,\cdots, 10\}$ \\ \hline
      MBR & $[0,5]$ MBps \\
      \hline
    \end{tabular}
    \label{tab:spaces}
\end{table}

Our goal is to adapt one policy trained on a service for other services or across changed radio environments. A noise is also simulated in the emulator by running another service in the background with random changes to its priority and MBR values. This forces the policies to constantly adjust the optimal control parameters and prevents a naive greedy approach from becoming optimal. 

\section{Results and Discussions} \label{sec:results}
\noindent This section presents the results of the following experiments.
\begin{enumerate}
    \item {\em Domain adaptation across services}. Here we train in one service and observe how it adapts to other services. We consider all pairs between URLLC, MIoT, and CV.
    
    \item {\em Domain adaptation across two radio environments}. Here we consider differing UE distribution across gnodeBs. We have used the URLLC service as the test bed.
\end{enumerate}

\subsection{Domain Adaptation across Service Types}
Here we train a RL policy on each source domain. For the CV domain, the goal is to maintain QoE at a target value, whereas, for the URLLC and mIoT domains, the goal is to maintain {\em PER} at a target value. In other words, QoE is the controlled variable for CV, and PER is the controlled variable for URLLC and mIoT. The action space for all these domains is the same, that is, the priority of the service and MBR of UE's within the service. For all the domains, the observation space considered is: 
\begin{align}
    \text{Observation Space} = \langle K, |K-target|/K \rangle
\end{align}
where $K$ is the current value of the controlled variable measured and $target$ is the target KPI value. Such a representation of state space provides two advantages: ($1$) It helps in creating a {\em goal-conditioned learning}, which enables the algorithm to achieve any value of desired goal in the state space, ($2$) This enforces a pseudo-normalization across the state space given that QoE and PER are measured in slightly different scales.

\noindent The action space is  
\begin{align}
    \text{Action Space} = [\text{Prio}, \text{MBR}_1, \text{MBR}_2, \cdots, \text{MBR}_N]
\end{align}
where MBR$_i$ is the value of MBR chosen for the $i^{th}$ UE, $N$ is the number of UEs within the service, and Prio is the priority value chosen for the service. In all these scenarios, we assume $N=4$ within each service. 

\noindent The reward considered is 
\begin{align}
    \text{Reward} = -|K-target|
\end{align}
where $K$ is the current value of the controlled variable. 

For each service type, we train a single-agent RL policy to achieve the maximum global reward. Now, the goal is to adapt the policy learned to a target domain, namely one of the other two service types, using adversarial training. To estimate the goodness of the adapted policy, we use the RL policy trained on the target domain as the golden reference. We also determine the number of samples required for the adapted policy to reach a 98\% success rate, that is, $2\%$ deviation with respect to the golden target policy. 

We use the domain adaptation method discussed in Section~\ref{sec:domain_adapt} to adapt each source policy on the other two target domains. Essentially, we prepare the following policies for comparison:  
\begin{enumerate}

    \item {\em Learn a policy from scratch for the target domain}. We collect samples from the target domain and pause the training after every 100 samples. During each pause, we evaluate the trained model for $8$ episodes. For this set of episodes, we compute the average deviation with respect to the pre-trained golden policy for the target domain, and also the standard deviation of the deviation. We continue the above process until the average deviation reaches $2\%$ in the test episodes. The intention of this exercise is to study the progression of learning for a policy trained from scratch. It may be noted that this exercise is for comparison with domain adaptation, and for a new service, a golden policy does not exist a priori.

    \item {\em Learn a policy through adaptation from a source domain}. Here we use the method discussed in Section~\ref{sec:domain_adapt} to adapt a source policy on the target domain. We initialize the correspondence models with random weights and measure the performance of the source policy with these correspondence models for $8$ episodes. Further, we collect $100$ samples from both the source domain and target domain and update correspondence models and measure the performance for $8$ episodes on the target domain. This process is repeated for every $100$ samples until the average deviation reaches $2\%$ in the test episodes.
    
\end{enumerate}
We now outline the results with each of URLLC, mIoT, and CV as source domains.

\subsubsection{Scenario-1: URLLC as source domain}
First, we train an RL policy to maintain the QoE at the target. It is to be noted that the goal is to be on the target and not exceed it by any arbitrary amount. In the presence of random noise, such a goal helps to arrive at a policy that is not trivially greedy in resource accumulation. The performance of the policy on the URLLC service is shown in Figure~\ref{fig:URLLC_exec}. From the plot, it is clear that the policy performs quite well when tested on URLLC service.

Before pursuing domain adaption, let us check the performance of the URLLC policy on mIoT and CV service without any adaptation. The performance plots are shown in Figure~\ref{fig:URLLC_exec_m_c}. From the plots, it can be seen that the performance of the URLLC source policy is not satisfactory on either of the target domains. This highlights the need for domain adaptation.

\begin{figure*}[ht]
    \centering
    \centering
     \subfloat[mIoT]{\includegraphics[scale=0.4]{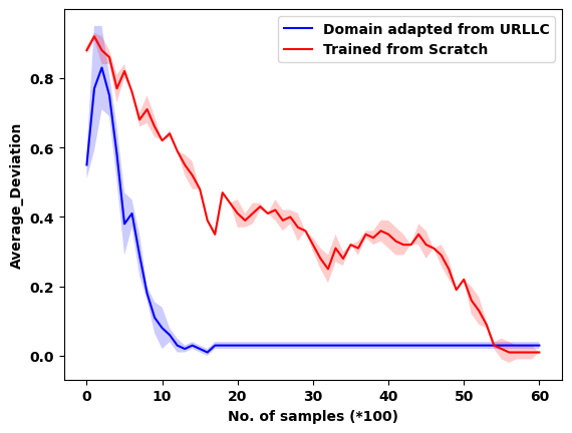}\label{fig:URLLC_miot} $\qquad$}
     \subfloat[CV]{\includegraphics[scale=0.4]{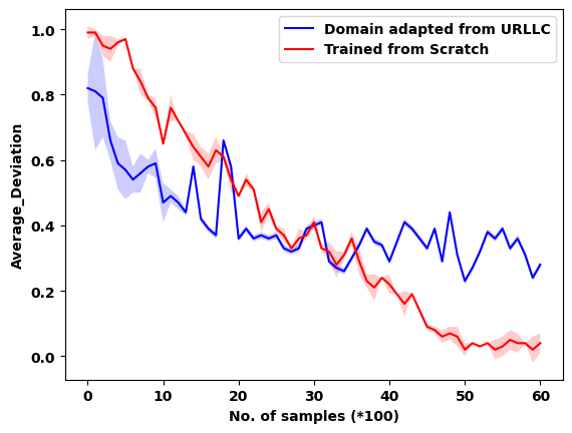}\label{fig:URLLC_CV}}
    \caption{Comparison of domain adapted policies from URLLC as source with policies learnt from scratch}
    \label{fig:my_label}
\end{figure*}

\begin{figure*}[ht]
    \centering
    \centering
     \subfloat[URLLC]{\includegraphics[scale=0.4]{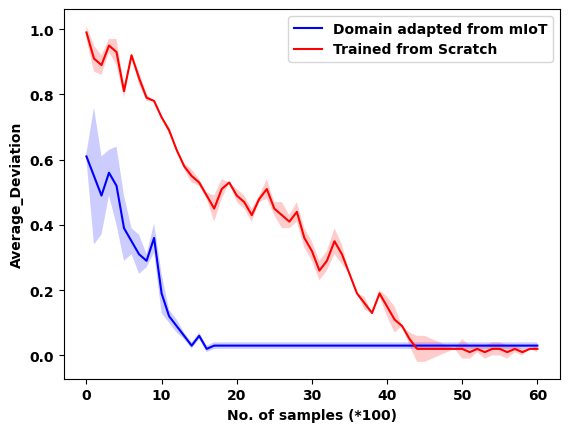}\label{fig:miot_urllc} $\qquad$}
     \subfloat[CV]{\includegraphics[scale=0.4]{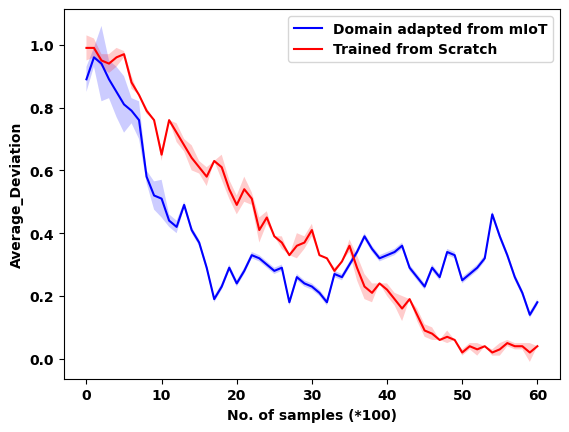}\label{fig:miot_CV}}
    \caption{Comparison of domain adapted policies from mIoT as source with policies learnt from scratch}
\end{figure*}

\begin{figure*}[ht]
    \centering
    \centering
     \subfloat[URLLC]{\includegraphics[scale=0.4]{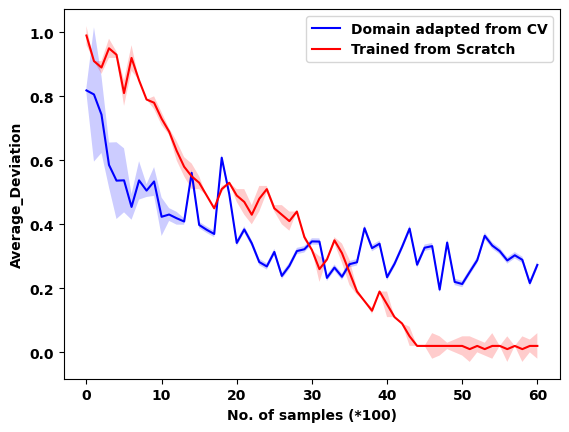}\label{fig:CV_urllc} $\qquad$}
     \subfloat[mIoT]{\includegraphics[scale=0.4]{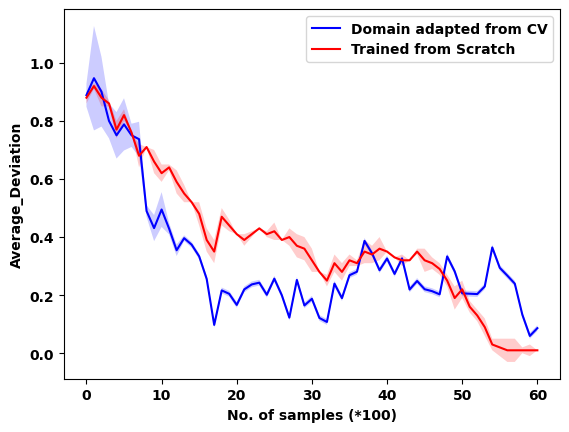}\label{fig:CV_miot}}
    \caption{Comparison of domain adapted policies from CV as source with policies learnt from scratch}
\end{figure*}

The performance of the policies for mIoT and CV adapted from the source URLLC policies are compared with the corresponding policies learnt from scratch, and the results are presented in Fig~\ref{fig:URLLC_miot} and Fig~\ref{fig:URLLC_CV} respectively. From Figure~\ref{fig:URLLC_miot}, on the mIoT target domain, we observe that the domain adapted policy requires about $1200$ samples to reach the target $2\%$ average deviation, whereas the policy learned from scratch requires around $5500$ samples to reach the same target. This shows the sample efficiency of the domain-adaptation approach, which is a key to rapid adaptation. Interestingly, such an adaptation using Cycle-GAN generated correspondences do not require the reward structure for the new environment.

Figure~\ref{fig:URLLC_CV} shows that for the CV domain, the policy trained from scratch requires around $5000$ samples to reach the target $2\%$ average deviation. On the other hand, the domain-adapted policy from the source domain does not reach the desired goal even with more samples, though initially it starts learning much faster than the policy trained from scratch and ends up improving the performance over the source policy. Hence here the domain-adaptation does not yield a near-optimal performance, unlike the case of mIoT.

A possible explanation of why the domain-adaption worked on the URLLC to mIoT domain, but did not work on the URLLC to CV domain is given herein. Usually, in telecom networks, the traffic pattern for each service is different based on what type of data they transmit, and how they transmit. In the case of URLLC and mIoT, we can see a similar traffic pattern whereas, in CV, we can see totally different patterns of data. Various domain experts also confirm the same on the similarity or dissimilarity between the three services. This leads us to infer that if domains are similar, domain adaptation will give better performance and would be significantly more sample efficient as compared to training from scratch. On the other hand, if the domains are less similar, domain adaptation can require a higher number of samples, and may not be comparable to training from scratch. This motivates us to use the heuristic presented in Section~\ref{sec:heuristic}, which notably points out the proximity between URLLC and mIoT, as well as the lack of proximity between CV and these services.

\subsubsection{Scenario-2: mIoT as source domain}
Here we explore the adaptation of a mIoT source policy to URLLC and CV. Figure~\ref{fig:miot_urllc} compares the learning of a URLLC policy from scratch with the adaptation of the source mIoT policy to URLLC. From the plot, it is evident that arriving at a domain-adapted policy requires less number of samples when compared to that of a policy trained from scratch. This is in agreement with our heuristic, which predicts the proximity between URLLC and mIoT.

On the other hand, Figure~\ref{fig:miot_CV} shows that adapting a mIoT source policy to CV is not a good option, as the domain-adapted policy does not show good performance even when correspondence models were trained with a large number of samples. Again this is in agreement with our heuristic, which estimates the lack of proximity between mIoT and CV.
 
\subsubsection{Scenario-3: CV as source domain}
Here we study the case where CV is the source domain, and the target domains are URLLC and mIoT. The results are shown in Fig~\ref{fig:CV_urllc} and Fig~\ref{fig:CV_miot} respectively. From the plots, it is evident that both domain adaptations resulted in poor performance when compared with policies trained from scratch. This is also in agreement with our heuristic, which shows CV to be distant from both URLLC and mIoT.

Our results support the proposed heuristic for estimating service proximity and establish the benefit of domain adaptation from a proximal service policy. Domain adaptation is sample efficient for proximal services as compared to training a new policy from scratch. Moreover, the entire methodology is completely automated, as reward signals are not needed in the new environments. We believe that domain adaption will be particularly beneficial in 6G scenarios, where we will have a multitude of services, some of them dynamically defined. Every time a new service comes, the system can automatically adapt from a proximal service on-the-fly, instead of learning from scratch. 

\subsection{Domain Adaptation across different Radio Environments}
This section highlights the benefit of domain adaptation when the underlying radio environment changes. Usually, the mobility pattern in a city is quite dynamic, and as UEs move, the traffic distribution across gNodeBs change. When the change in distribution is significant, the policy needs to adapt.

\begin{figure}
    \centering
     \subfloat[Existing distribution (Gaussian)]{\includegraphics[scale=0.35]{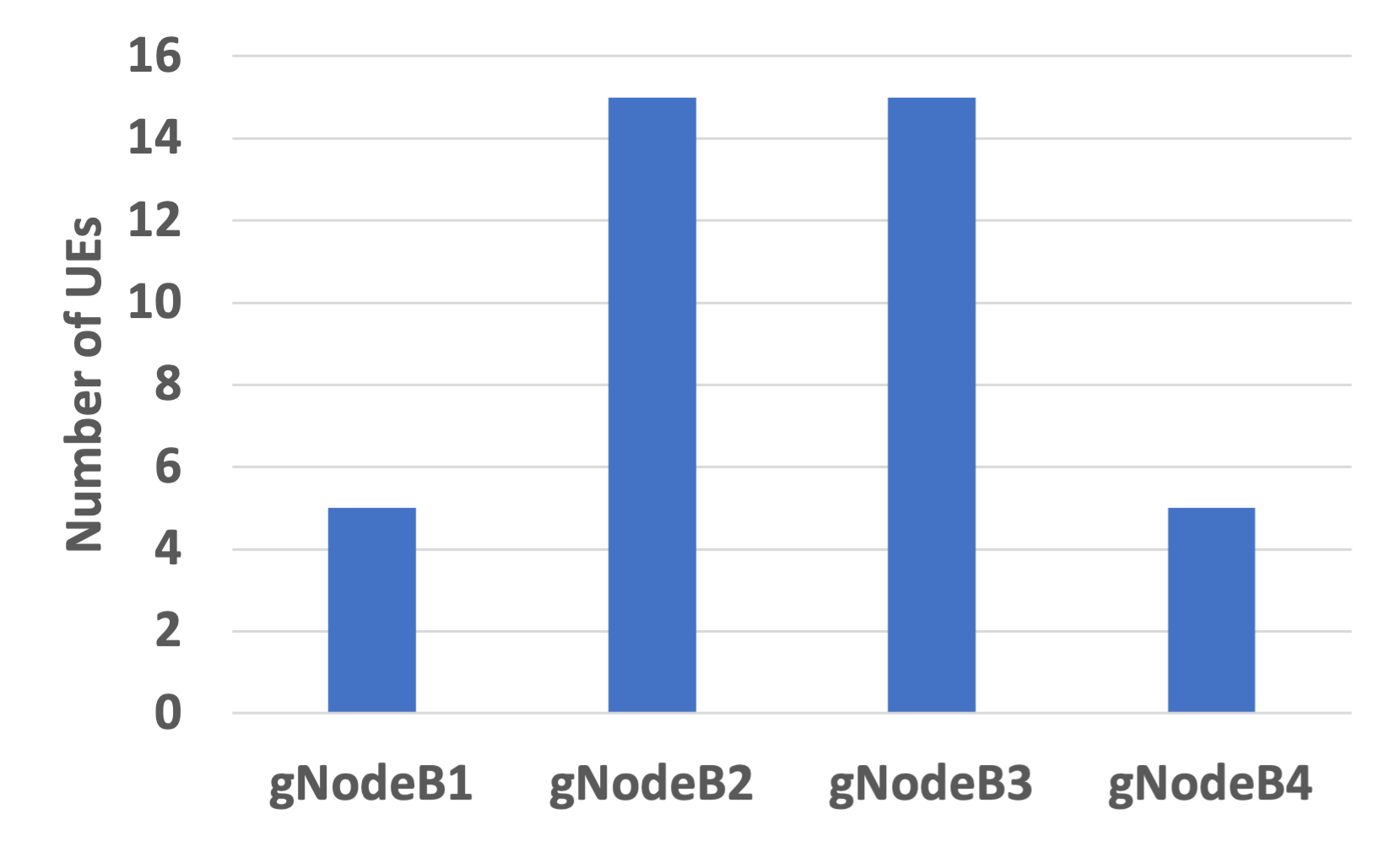}\label{fig:Gauss_urllc}}
     \subfloat[Modified distribution (Uniform)]{\includegraphics[scale=0.35]{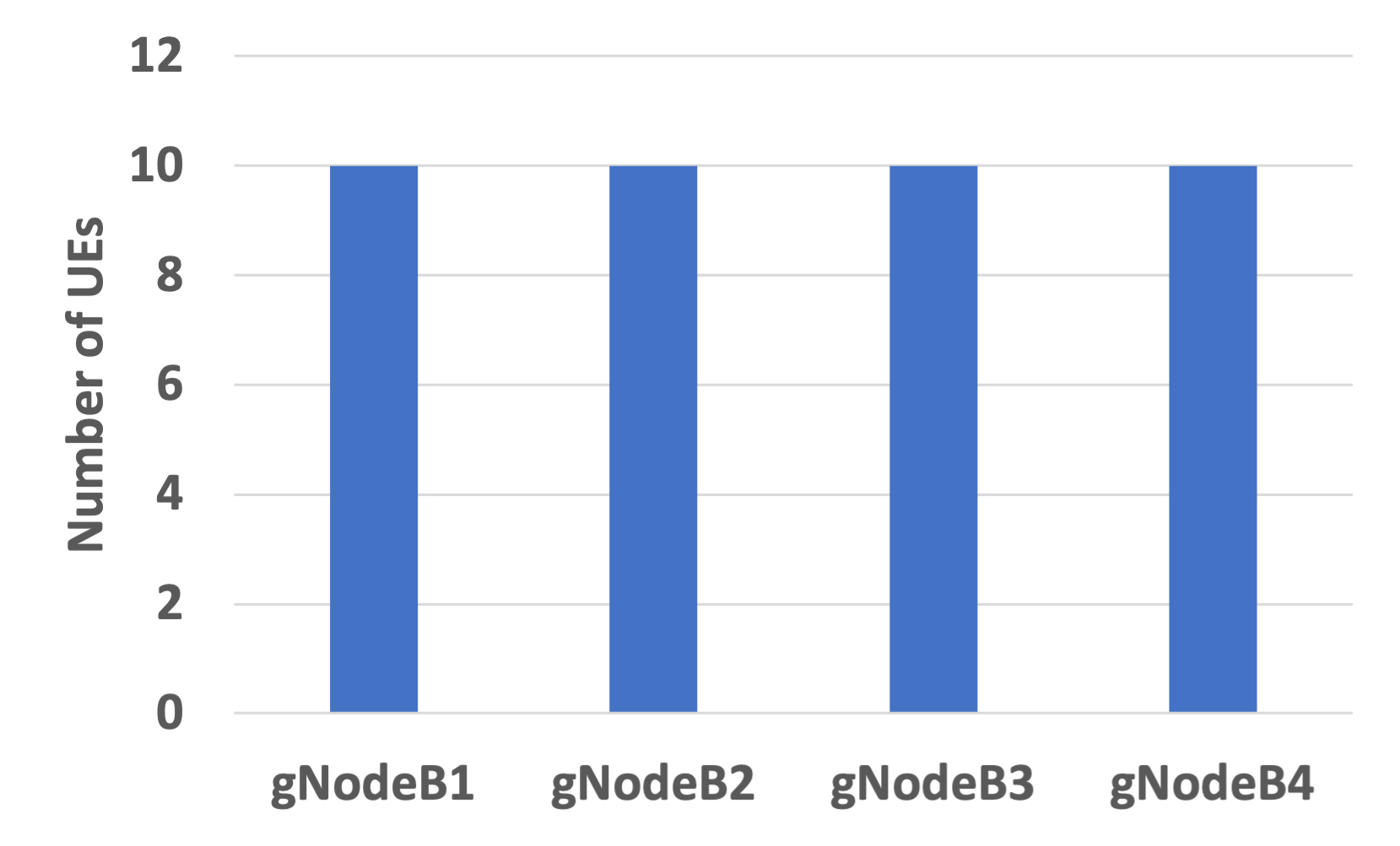}\label{fig:Unif_urllc}}
    \caption{Distribution of UEs with respect to gNodeBs}
    \label{fig:modified_distr_1}
\end{figure}

\begin{figure*}
    \centering
    \subfloat[Gaussian to Uniform]{\includegraphics[scale=0.4]{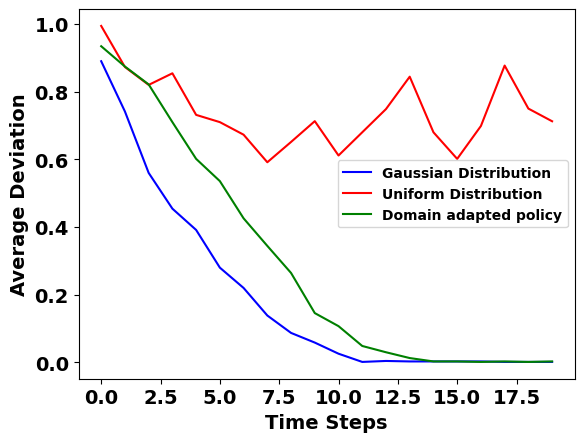}\label{fig:gauss_unif_exec} $\qquad$}
    \subfloat[Uniform to Poisson]{\includegraphics[scale=0.4]{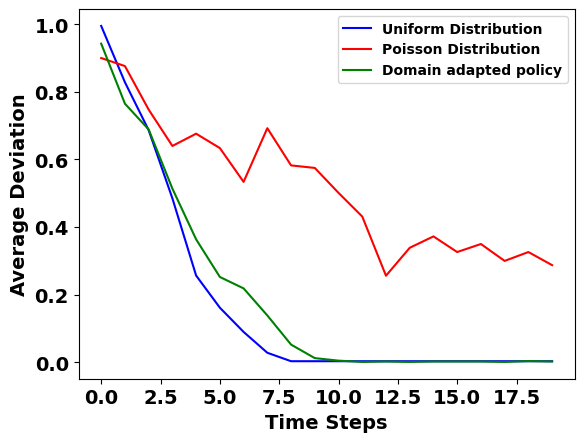}\label{fig:unif_pois_exec}}
    \caption{Comparison of execution of source policy, on source(Blue) and new distribution (Red), and with domain-adapted source policy on new distribution (Green). A change of distribution simulates the change in the underlying radio environment. } \label{fig:domainshift}
\end{figure*}

As an example, Fig~\ref{fig:modified_distr_1} shows a change in the distribution of UEs across gNodeBs from Gaussian to Uniform. Figure~\ref{fig:gauss_unif_exec} compares the performance of the source policy on both environments with the domain-adapted policy, which learns to adapt from Gaussian to Uniform. From the plot, it is evident that the performance of the original policy is sub-optimal on the new distribution, but the domain-adapted policy does much better. Given that such situations happen often in radio environments, frequent retraining is typically infeasible, and domain-adaptation seems to be the way forward.

\begin{figure}
    \centering
     \subfloat[Existing distribution (Uniform)]{\includegraphics[scale=0.35]{Uniform_Distribution.png}\label{fig:Unif_urllc_1}}
     \subfloat[Modified distribution (Poisson)]{\includegraphics[scale=0.35]{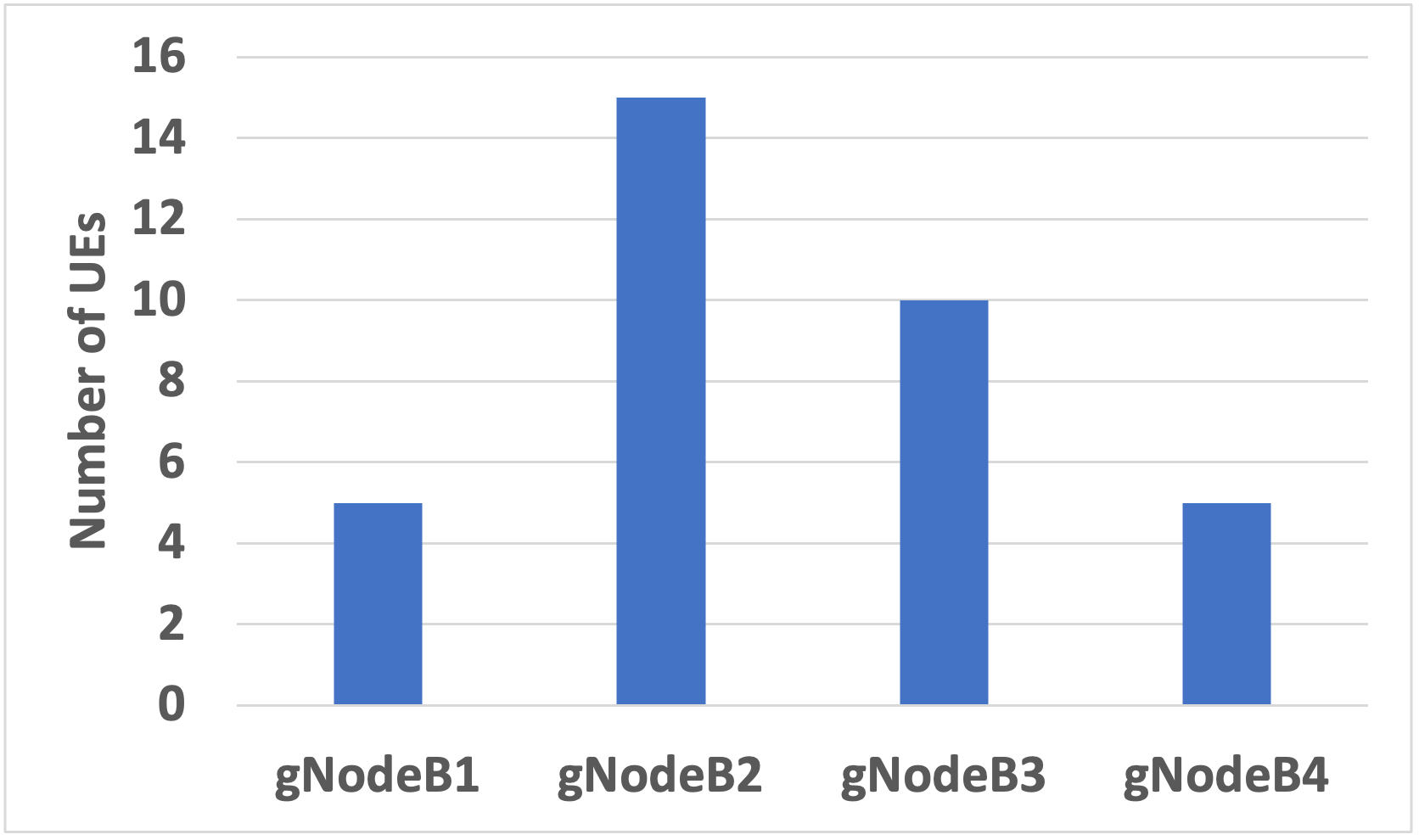}\label{fig:poi_urllc}}
    \caption{Change in distribution of UEs w.r.t gNodeBs}
    \label{fig:modified_distr_2}
\end{figure}

\begin{table}
    \centering
    \caption{Samples required to achieve $98\%$ performance}
    \begin{tabular}{lcc}
     \textbf{\sf Scenario}    & \textbf{\sf Domain adaptation} & \textbf{\sf  Trained from scratch} \\
     \hline \hline
       Gaussian to Uniform  & $600$ & $5000$ \\
       \hline 
       Uniform to Poisson  & $500$ & $4600$ \\
       \hline
    \end{tabular}
    \label{tab:dist_training}
\end{table}

Figure~\ref{fig:unif_pois_exec} presents our results when the distribution shifts from Uniform to Poisson, as in Fig~\ref{fig:modified_distr_2}. Again the domain-adapted policy performs well in the modified distribution. 

Table~\ref{tab:dist_training} compares the training performance of the domain-adapted policies with those learned from scratch. The domain-adapted policy requires far fewer samples to converge to less than $2\%$ deviation as compared to the policies learned from scratch. This enables quick and automated adaptation without reward signals in new environment. The reduced sample complexity and computational requirement are significant, motivating widespread adoption of the same in edge devices.

\section{Conclusions}\label{sec:conc}
In this paper, we propose domain adaptation approaches to accommodate new telecom services and changes in radio environments. Since the 6G environment is expected to contain many services, and some of them may evolve dynamically, the proposed method advocates an approach for learning from already trained services to prepare an adapted policy of acceptable performance in a quick time. The approach uses a combination of Cycle GANs to construct the correspondence networks between old services and new services instead of learning a policy on new services. This method does not need reward information or transition dynamics of the new domain.  Also, we introduce a heuristic for deciding which of the existing policies may be reused for a successful domain adaptation. Experimental results on the custom-built emulator show the efficiency of the proposed approach when compared with training from scratch on new services.

\bibliographystyle{plain}
\bibliography{References}
\end{document}